\documentclass[runningheads]{llncs}
\usepackage[vlined,ruled,linesnumbered]{algorithm2e}
\usepackage{setspace}
\usepackage{amsmath,amssymb,amsfonts}
\usepackage{textcomp}
\usepackage{changes}
\usepackage{lipsum}
\usepackage{wrapfig}
\usepackage{pgfplots}
\pgfplotsset{compat=1.17}
\usepackage{orcidlink}
\usepackage{cleveref}

\crefname{figure}{Fig.}{Figs.}
\Crefname{figure}{Figure}{Figures}
\crefname{section}{Sect.}{Sects.}
\Crefname{section}{Section}{Sections}
\crefname{algorithm}{Alg.}{Algs.}
\Crefname{algorithm}{Algorithm}{Algorithms}
\crefname{table}{Table}{Tables}

\usepackage{array}  % used for related work table
\usepackage[misc]{ifsym}    % used for letter symbol
\usepackage{makecell}

\renewcommand{\textapprox}{\raisebox{0.5ex}{\texttildelow}}

\begin{document}
\renewcommand{\floatpagefraction}{0.8}%minimum fraction of floatpage that should have floats

\title{A Rule-Based Behaviour Planner for Autonomous Driving}

\author{
Fr\'{e}d\'{e}ric Bouchard\,\textsuperscript{\Letter}\,\orcidlink{0000-0002-9529-7577}\and Sean Sedwards\,\orcidlink{0000-0002-2903-0823} \and Krzysztof Czarnecki\,\orcidlink{0000-0003-1642-1101}}

\institute{University of Waterloo, Canada
\email{\{frederic.bouchard,sean.sedwards,krzysztof.czarnecki\}@uwaterloo.ca}}

\authorrunning{Fr\'{e}d\'{e}ric Bouchard, Sean Sedwards, and Krzysztof Czarnecki}

\maketitle

\begin{abstract}

Autonomous vehicles require highly sophisticated decision-making to determine their motion. This paper describes how such functionality can be achieved with a practical \emph{rule engine} learned from expert driving decisions. We propose an algorithm to create and maintain a rule-based behaviour planner, using a two-layer \emph{rule-based theory}. The first layer determines a set of feasible parametrized behaviours, given the perceived state of the environment. From these, a resolution function chooses the most conservative high-level maneuver. The second layer then reconciles the parameters into a single behaviour. To demonstrate the practicality of our approach, we report results of its implementation in a level-3 autonomous vehicle and its field test in an urban environment.

\end{abstract}

\section{Introduction}

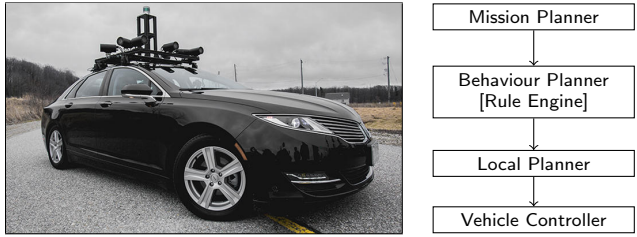
\begin{figure}[t!]
  \vspace{0.5em}
  \centering
    \frame{\includegraphics[trim={5.5cm 0 5.5cm 0 },clip,height=8.7em]{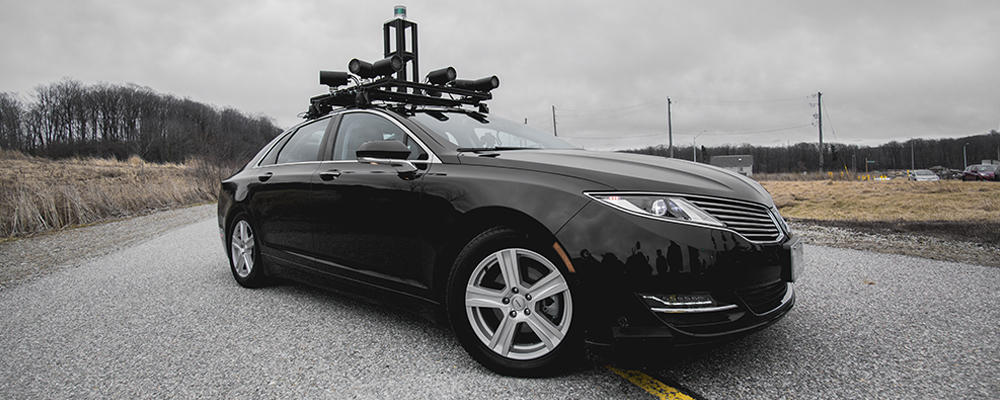}}
    \hspace{0.5em}
    \begin{tikzpicture}
    \sf\scriptsize
    \draw[every node/.style={draw},text width=2.5cm,align=center]
    (0,0)node(MP){Mission Planner}
    (0,-1.0)node(BP){Behaviour Planner [Rule Engine]}
    (0,-1.95)node(LP){Local Planner}
    (0,-2.7)node(VC){Vehicle Controller};
    \draw[->](MP)edge(BP)(BP)edge(LP)(LP)edge(VC);
    \end{tikzpicture}
    \caption{Motion planning architecture of our autonomous vehicle}
    \label{fig:moose-motion-planner}
    \vspace{-5pt}
\end{figure}

The motion planning problem in autonomous vehicles is computationally challenging~\cite{motion-planning} and is typically decomposed into three sub-problems~\cite{paden2016survey}: (i) mission planning; (ii) behaviour planning; and (iii) local planning.
This structure is depicted on the right of~\cref{fig:moose-motion-planner}.
In our autonomous vehicle, the mission planner receives starting and target locations, and determines the sequence of lanes on which the autonomous vehicle must drive. This sequence is converted into intents (e.g. turning right at the next intersection) and is sent to the behaviour planner, along with the environment representation. The behaviour planner then generates a sequence of high-level parametrized driving maneuvers to navigate through the environment towards the specified goal. The local planner finds a smooth trajectory that meets the required behaviour and comfort. Finally, the trajectory is used by the vehicle controller to determine the steering, throttle, and braking commands.

Early approaches to behaviour planning used finite state machines~\cite{montemerlo2008junior,urmson2009autonomous}.
Such systems are typically difficult to maintain because of the inherent complexity of the driving problem. Combinations of state machines, decomposing the problem into sub-problems, can mitigate this lack of maintainability~\cite{shekhar1994use}. The resulting hierarchy of state machines often introduces the need of precedence tables~\cite{niehaus1991expert}, which is a concept that is also familiar to rule-based systems~\cite{furnkranz2012foundations}.

Recently, there has been a strong trend to use deep learning for autonomous driving tasks. End-to-end machine learning approaches have been shown to handle basic driving tasks~\cite{bojarski2016end}, while imitation learning of behaviours~\cite{ChauffeurNet} or trajectories~\cite{kuderer2015learning} can produce highly nuanced behaviours in complex road environments.
The success of deep learning is in part due to its ability to learn the structure of a problem at the same time as solving it~\cite{deSainteMarie2021}.
Its main drawback is that the resulting policies lack transparency and explainability~\cite{lipton2018mythos}.

By contrast, rule-based systems have the advantage of structured knowledge, greatly aiding explainability, safety and trust.
Early works proved the concept of sophisticated direct vehicle control using rules~\cite{niehaus1991expert,zimmerman2004implementing}, while more recent work has tended to use rule-based systems indirectly, such as to validate, bound or improve driving policies~\cite{yay2014using,vanderhaegen2016rule}.
Finding the structure of a  problem remains an important challenge for rule-based systems~\cite{Furnkranz2020,deSainteMarie2021}, as is the maintenance and update of the rule base as the problem evolves.
To mitigate these challenges, hybrid approaches distil rules from deep-leaned policies~\cite{Fu2022} or use pre-defined rules to constrain the action space of deep learning~\cite{Lee2019}.

In this paper, we define a two-layered rule engine for behaviour planning and present an algorithm to create and maintain its rule-based theory.
Rather than learn the structure, we decompose the problem into meaningful concepts, to maximize explainability.
Our maintenance algorithm exploits this and allows a theory to be learned incrementally from a set of expert-provided examples that may be augmented over time.
We have designed the layers to be modular, allowing us in future to add layers and construct ``deep theories'' when we extend the operational design domain (ODD) of our vehicle.

To demonstrate the success and practicality of our approach, we have used it to construct a prototype implementation that we have deployed in our autonomous vehicle (\cref{fig:moose-motion-planner}).
We report the results of our vehicle's 110\,km field test in a busy urban environment, during which our rule engine was able to make decisions at up to 300\,Hz and achieve a similar level of autonomy (98\%) to a highly-cited deep-learning approach~\cite{bojarski2016end}. While we acknowledge that this comparison is not rigorous, because the tasks and ODDs are not rigorously aligned, we nevertheless claim that our approach has the advantage of being able to immediately identify and potentially fix any logical errors of its decisions.

While there is much in the literature that may be considered related to our work, there appears to be very little that is directly comparable in terms of the levels of detail, implementation and practical achievement that we present.
The following table summarizes our approach in the context of selected works that apply rule-based decision-making to autonomous driving.

\begin{center}
\setlength{\tabcolsep}{0.1cm} % for the horizontal padding
\sf\small
\begin{tabular}{>{\raggedright}p{0.7cm}|>{\centering}p{1.5cm}|>{\centering}p{1.5cm}|>{\centering}p{1.5cm}||>{\centering}p{1.5cm}|>{\centering}p{1.5cm}|>{\centering\arraybackslash}p{1.5cm}}
\hline
 \makecell[lt]{\\Ref.} & Explicit Layers & Unordered Rule Base & Learning Algorithm & Urban ODD & Realistic Perception & Validated in Reality\\
 \hline
 Ours & \checkmark & \checkmark & \checkmark & \checkmark & \checkmark & \checkmark\\
 \cite{vanderhaegen2016rule} & implicit & \checkmark & \checkmark & & & \\
 \cite{Zhao2016} & implicit & \checkmark & & subset & \checkmark & \checkmark \\
 \cite{Wang2019} & implicit & \checkmark & & & \checkmark & \\
 \cite{Kapania2019} & implicit & & & subset & & \\
 \cite{zimmerman2004implementing} & implicit & & & subset & & \\
 \cite{niehaus1991expert} & implicit & & & & & \\
 \cite{Fu2022} & & \checkmark & & \checkmark & \checkmark & \\
 \cite{Likmeta2020} & & \checkmark & & \checkmark & & \\
 \cite{Xiao2021} & & \checkmark & & subset & & \\
 \cite{Aksjonov2021} & & unspecified & & subset & & \\
 \cite{Lee2019} & & & & subset & & \\
 
 \hline
\end{tabular}
\end{center}

In~\cref{sec:rule-engine} we describe the conceptual data structures, rules and functions of the layers that comprise our rule engine, explaining how a single behaviour is chosen from a set of feasible options.
In~\cref{sec:learning} we define a backward-chaining coverage function and present an algorithm that uses it to learn and maintain the rule-based theory of the rule engine.
We also outline a knowledge engineering cycle that makes use of our algorithm to incrementally build a set of rules.
In~\cref{sec:results} we present the practical results of using our prototype rule engine, emphasizing its successful deployment in an extended drive on public urban roads.
We conclude in~\cref{sec:conclusion}, highlighting challenges and ongoing work.
\section{Rule Engine}
\label{sec:rule-engine}
In this section, we describe the two-layer rule-based theory that is the conceptual basis of our rule engine, noting that the syntactic sugar for expressing rules compactly, such as bounded quantifiers, and the many optimizations of our implementation are omitted to simplify our exposition. 

Each layer of the theory uses a set of unordered ``\textsf{IF} \textit{antecedent} \textsf{THEN} \textit{consequent}'' rules that map a set of input properties to a set of parametrized output behaviours. The first layer, called the \emph{maneuver layer}, takes properties of the external environment as input and outputs a candidate set of parametrized behaviours, which are then filtered according to their conservativeness. The resulting behaviours, which now share the same high-level maneuver, are then transformed to become the input of the second layer, called the \emph{parameter layer}. The parameter layer resolves the different parameters and outputs a single high-level maneuver with its parameter.

\subsection{Layers and Rules}

\sloppy Each layer of our rule-based theory is described by a tuple of finite sets $(\mathcal{O, A, V, P', C, R})$.
$\mathcal{O}$ is a set of objects recognized by the layer.
$\mathcal{A}$ is a set of attributes that an object may have (colour, speed, etc.)
$\mathcal{V}$ is a set of values that object attributes may take (green, 2.7, \textit{True}, etc.)
Triplets of type $(\mathcal{O,A,V})$ constitute \emph{input properties}.
For notational convenience, we write $\mathcal{O_A}$ for $\mathcal{O}\times\mathcal{A}$ and express properties in the forms $(\mathcal{O_A,V})$ and $\mathcal{O_A} := \mathcal{V}$. We call elements of $\mathcal{O_A}$ \emph{features}.
We also include in $\mathcal{V}$ the special value $\textit{undefined}$ that may be assigned to any feature that is not defined.
We say a property or feature is undefined if its value is $\textit{undefined}$.
$\mathcal{P'}$ is the set of \emph{output properties}, defined analogously to input properties, but w.r.t. the objects and attributes of the subsequent layer. This allows the output of one layer to become the input of the next layer.

$\mathcal{C}$\ is a set of logical constraints over features, which evaluate to either \textit{True} or \textit{False} and have the type $(\mathcal{O_A}, \{=, \leq, \geq \}, \mathcal{V})$ or $(\mathcal{O_A} ,\{=,\leq,\geq \}, \mathcal{O_A})$, with the obvious mathematical meaning.
E.g., $\textit{Ego}_\textit{Approaching} = \textit{Intersection}$, $\textit{Ego}_\textit{Speed} \geq \textit{LeadingVehicle}_\textit{Speed}$.
We also include in $\mathcal{C}$ the trivial constraint \textit{True} and note that the operators $\{\leq, \geq \}$ return \textit{False} whenever they encounter an \textit{undefined} property.

Each layer has an associated set of rules $\mathcal{R}$, in which a rule is a tuple of type $\left( \mathbb{P}\left(\mathcal{C}\right), \mathcal{B} \right)$. The first element, referred to as the rule's \emph{antecedent}, is a conjunction of constraints ($\mathbb{P}$ denotes the power set), and the second element, referred as the rule's \emph{consequent}, is the behaviour induced when the antecedent evaluates to true.
A behaviour $b\in\mathcal{B}$ has type $\left( \mathcal{H}, \mathbb{P}\left( \mathcal{P'} \right) \right)$, in which 
$\mathcal{H} = \{ \textit{Emergency-Stop},\textit{Stop},\textit{Yield}, \textit{Decelerate-}\linebreak\textit{To-Halt},\textit{Pass-Obstacle},\textit{Follow-Leader},\textit{Track-Speed}\}$
is the globally-defined set of high-level maneuvers we use in this work.
We refer to the second element of the tuple as the behaviour's \emph{parameter}. 
The syntax of rule antecedents is given by the following simple BNF grammar, in which a \textit{constraint} is $c \in \mathcal{C}$:
\begin{align*}
    \langle\textit{antecedent}\rangle &::= \langle\textit{antecedent}\rangle \textsf{ AND } \langle\textit{antecedent}\rangle
    \mid\textit{constraint}\nonumber
\end{align*}
The antecedent of a rule typically contains only a subset of available features, giving it only a partial view of the input, and thus capturing an abstract meaning.

We assume that the input to a layer is a complete set of properties that constitute a function $\mathcal{O_A}\to\mathcal{V}$.
We call this a \emph{scene} and denote by $\mathcal{S}$ the set of all scenes. A scene for the first layer contains  properties representing the road environment, whereas a scene for the second layer includes properties representing the candidate behaviours for the ego vehicle, generated by the first layer. 

A rule $r$ is then implicitly represented by a corresponding function $\mathcal{F}_r: \mathcal{S} \to \mathcal{B}$ that maps a scene $e \in\mathcal{S}$ to the behaviour $b\in\mathcal{B}$ that is $r$'s consequent, or the empty set:
\begin{equation}
\mathcal{F}_r(e) := \left\{
\begin{array}{ll}
b & \text{if\ the\ rule's\ antecedent\ evaluates\ to\ true}\\
\varnothing & \text{otherwise}\\
\end{array} \right.
\label{eq:rule-function}
\end{equation}
Given a rule-based theory $\mathcal{R}$ (a set of rules for a given layer), we lift~\eqref{eq:rule-function} to a function $\mathcal{F_R}:\mathcal{S}\to \mathbb{P}\left(\mathcal{B}\right)$, where
\begin{equation}
\mathcal{F_R}(e\in\mathcal{S}) := \{\mathcal{F}_r(e) \mid r\in\mathcal{R}\}.
\label{eq:theory-function}
\end{equation}

\subsection{Resolving a Single Behaviour}
\label{SS:RefinementFunction}

The maneuver layer outputs all the behaviours that are at least partially compatible with the perceived outside world, according to a set of rules denoted $\mathcal{R}_\textit{man}$ and corresponding function $\mathcal{F_R}_\textit{man}$ defined by~\eqref{eq:theory-function}.
The following is an example of a rule in  $\mathcal{R}_\textit{man}$, where we expect the ego vehicle to stop at the stop line when it approaches an intersection regulated by a stop sign:

\begin{center}
\small
\renewcommand{\arraystretch}{1.4}
\begin{tabular}{rl}
    {\sf IF} &  $\textit{Ego}_\textit{Approaching} = \textit{Intersection}$
       {\sf AND}  $\textit{Road}_\textit{HasStopLine} = \textit{True}$\\
    {\sf THEN} &  (\textit{Decelerate-To-Halt}, \{\textit{Stop\textsubscript{AtStopLine}} := \textit{True}\})
\end{tabular}
\end{center}

The output of the maneuver layer will often contain behaviours with incompatible high-level maneuvers, i.e., with different elements of $\mathcal{H}$, as well as behaviours having the same high-level maneuver, but with different parameters.
To eventually arrive at a single behaviour, we first narrow the range of behaviours seen in the output of the maneuver layer, using a relation $\succ$ that defines a total order over the conservativeness of high-level maneuvers.
We can thus write $\textit{Emergency-Stop}\succ\textit{Track-Speed}$ to mean \textit{Emergency-Stop} is more conservative than \textit{Track-Speed}.
We then use the corresponding partial order relation $\succeq$ to define a resolution function
\begin{multline}
\lambda_\textit{man}(\mathcal{F_R}_\textit{man}(e\in\mathcal{S})):=\\
\{(h,p)\in\mathcal{F_R}_\textit{man}(e) \mid \forall(h',p')\in\mathcal{F_R}_\textit{man}(e), h \succeq h'\},
\label{eq:lambda-man}
\end{multline}
which returns the behaviours sharing the highest priority maneuver.

The output of $\lambda_\textit{man}$ is fed to the input of the parameter layer, following a transformation into a scene expected by the parameter layer, i.e., a function of type $(\mathcal{O_{A}})_\textit{par}\to\mathcal{V}_\textit{par}$, where $(\mathcal{O_A})_\textit{par}$ and $\mathcal{V}_\textit{par}$ are the features and values, respectively, of the parameter layer.
We thus define a transformation function comprising the union of three sets:
\begin{multline}
\label{eq:transformation-par}
\begin{array}{rl}
\mathcal{T}_\textit{par}(e\in\mathcal{S}) :=\quad & \{p\mid(h,p)\in\lambda_\textit{man}(\mathcal{F_R}_\textit{man}(e))\}\\
\cup & \{(\textit{Maneuver}_h,\textit{True})\mid(h,p)\in\lambda_\textit{man}(\mathcal{F_R}_\textit{man}(e))\}\\
\cup & \{(o_a,\textit{undefined})\mid o_a\in(\mathcal{O_A})_\textit{par} \setminus\\
\end{array}\\
\{(o_a)'\mid(h,((o_a)',v))\in\lambda_\textit{man}(\mathcal{F_R}_\textit{man}(e))\}
\}
\end{multline}
The first set contains all the parameters output by $\lambda_\textit{man}$, now interpreted as input properties of the parameter layer. The second set is a singleton containing a property that encodes the chosen high-level maneuver. The third set contains properties that map all the undefined features to the $\textit{undefined}$ value.

The output of $\mathcal{T}_\textit{par}$ corresponds to a single high-level maneuver with possibly ambiguous parameters. The purpose of the parameter layer is to resolve this ambiguity, using different properties and a different set of rules to the maneuver layer. We denote the parameter layer's rules and their corresponding function by $\mathcal{R_\textit{par}}$ and $\mathcal{F_R}_\textit{par}$, respectively.
The following example parameter rule guarantees that whenever \textit{Decelerate-To-Halt} does not target the stop-line, but instead targets the end of the lane, then this latter should be included in the final set of parameters:
\begin{center}
\small
\renewcommand{\arraystretch}{1.4}
\begin{tabular}{rll}
    \textsf{IF} & & \textit{Maneuver\textsubscript{Decelerate-To-Halt}} = \textit{True}\\
       & \textsf{AND} & \textit{Stop\textsubscript{AtStopLine}} = \textit{undefined}
       \textsf{AND} \textit{Stop\textsubscript{AtEndOfLane}} = \textit{True}\\
    \textsf{THEN} & & \big(\textit{Decelerate-To-Halt},\{\textit{Ego\textsubscript{StopAt}} := \textit{AtEndOfLane}\}\big)
\end{tabular}
\end{center}

The output of the maneuver layer is a set of behaviours with the same high-level maneuver and parameters that are consistent with one another. The resolution function for the parameter layer returns a single behaviour with its parameter being the union of the parameters returned by the parameter layer:
\begin{equation}
\lambda_\textit{par}\left(\mathcal{F}_{\mathcal{R}_\textit{par}}(\cdot)\right) :=
\left(h\mid\exists(h,p)\in\mathcal{F}_{\mathcal{R}_\textit{par}}(\cdot),\{p\mid(h,p)\in\mathcal{F}_{\mathcal{R}_\textit{par}}(\cdot)\}\right).\label{eq:lambda-par} 
\end{equation}

\begin{figure}
    \centering
    \begin{tikzpicture}
    \sf\footnotesize
    % grey boxes
    \draw[every node/.style={fill=black!10},minimum width=2.5cm,minimum height=4.1cm]
    (2,0.2)node{}
    (6,0.2)node{};
    % white data flow boxes
    \draw[draw=blue!50,fill=blue!10]
    (0.25,-1.4)rectangle(2.25,1.4)
    (2.75,-1)rectangle(6.25,1)
    (6.75,-0.5)rectangle(7.75,0.5);
    % layers, vertical labels
    \draw[every node/.style={draw,rotate=90,fill=white},minimum width=3.4cm, minimum height=0.5cm]
    (0,0)node{perceived state}
    (1.5,0)node{maneuver rules}
    (2.5,0)node{compatible  behaviours}
    (4,0)node{transform properties}
    (5.5,0)node{parameter rules}
    (6.5,0)node{single behaviour}
    (8,0)node{$\rightarrow\text{local planner}$};
    % horizontal labels
    \draw
    (0,2.5)node{Input}
    (1.5,2.0)node{$\mathcal{R}_\textit{man}$}
    (2,2.5)node{Maneuver Layer}
    (2.5,2.0)node{$\lambda_\textit{man}$}
    (4,2.0)node{$\mathcal{T}_\textit{par}$}
    (4,2.5)node{}
    (5.5,2.0)node{$\mathcal{R}_\textit{par}$}
    (6,2.5)node{Parameter Layer}
    (6.5,2.0)node{$\lambda_\textit{par}$}
    (8,2.0)node{}
    (8,2.5)node{Output};
    \end{tikzpicture}
    \caption{Diagrammatic representation of two-layer rule engine}
    \label{fig:RE}
\end{figure}
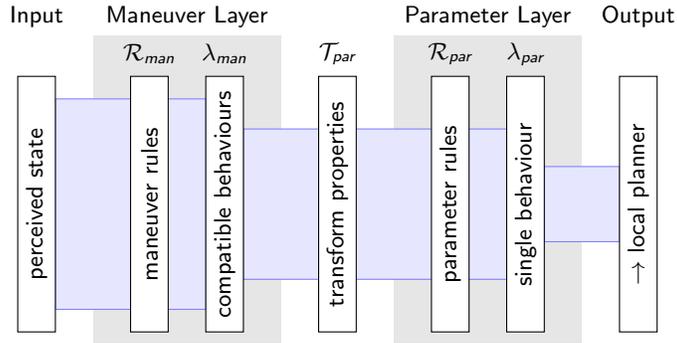

The overall function of our rule engine is thus given by
\begin{equation}
RE(e\in\mathcal{S}) := \lambda_\textit{par}\left(\mathcal{F}_{\mathcal{R}_\textit{par}}( \mathcal{T}_\textit{par}(e) )\right),
\nonumber
\end{equation}
which we also refer to as the driving policy.
\Cref{fig:RE} shows a diagrammatic representation of the rule engine: (i) sensors present a perceived  state to the maneuver layer, which identifies a set of compatible, conservative behaviours using $\mathcal{R}_\textit{man}$ and $\lambda_\textit{man}$; (ii) $\mathcal{T}_\textit{par}$ transforms and completes the resulting properties for input to the parameter layer; (iii) the parameter layer resolves a single behaviour using $\mathcal{R}_\textit{par}$ and $\lambda_\textit{par}$, which is sent to the local planner.

\subsection{Inference Example}

\begin{wrapfigure}{r}{0.5\textwidth}
    \centering
    \vspace{-2em}
    \includegraphics[width=0.5\textwidth,trim={0.5cm 0.5cm 3.5cm 0.5cm},clip]{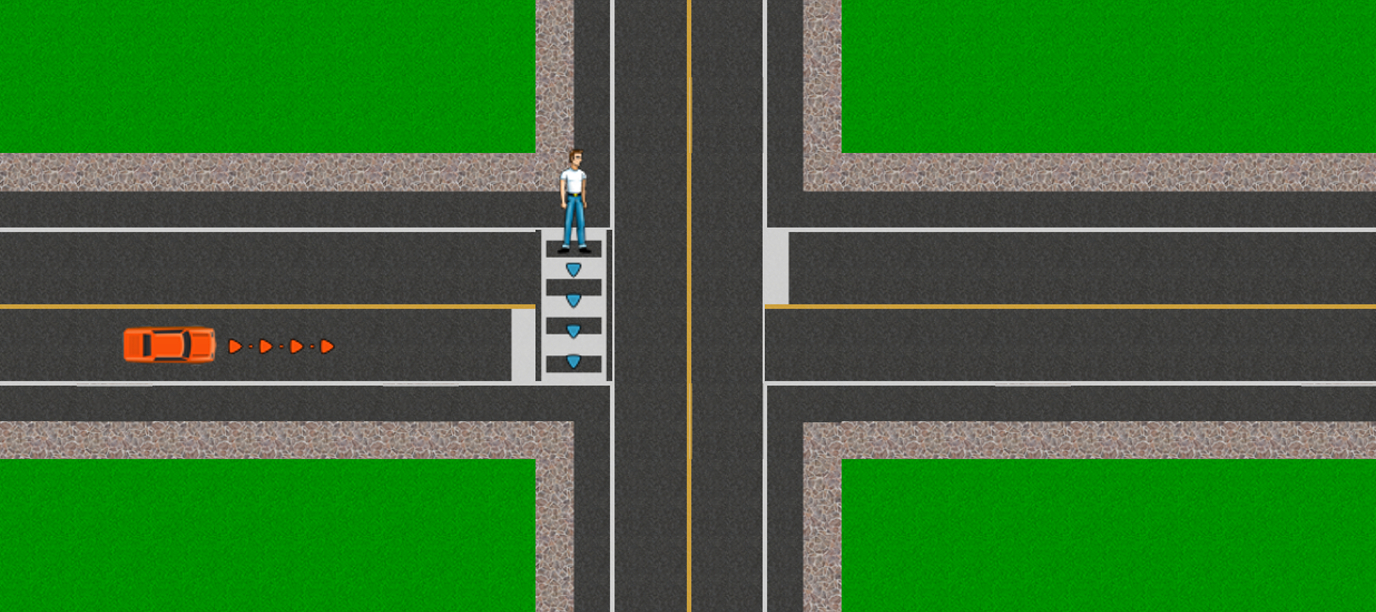}
    \caption{Example scene: autonomous vehicle approaches intersection with crosswalk}
    \label{fig:simple-scene}
    \vspace{-1em}
\end{wrapfigure}

To give an intuition of how the rule engine makes a decision, we present a toy example based on the scene illustrated in~\cref{fig:simple-scene}:
the autonomous (\textit{Ego}) vehicle approaches an intersection regulated by a stop line, while a pedestrian concurrently negotiates the crosswalk.
We exclude all elements of the rule engine not relevant to the scene, noting that this simple example is not intended to motivate the two-layered structure of the rule engine, which is required for the significantly greater complexity encountered in realistic applications.

The scene in~\cref{fig:simple-scene}, denoted $s$, is defined with a minimal set of features by
\begin{align*}
    s := \{
    & \textit{Ego\textsubscript{Approaching}} := \textit{Intersection}, 
    && \textit{Ego}_\textit{Speed} := 35\,\text{km/h},\\
    & \textit{Ego}_\textit{At} := \textit{undefined},
    && \textit{Road}_\textit{SpeedLimit} := 50\,\text{km/h},\\
    & \textit{Crosswalk\textsubscript{Obstructed}} := \textit{True},
    && \textit{Road}_\textit{HasStopLine} := \textit{True}
    \,\}.
\end{align*}

We define a set of \textit{Ego} behaviours relevant to $s$, having parameters specified w.r.t. the input features of the parameter layer:
\begin{align*}
    b_1 &:= (\textit{Track-Speed}, \{ \textit{Target}_\textit{Speed} := \textit{Road}_\textit{SpeedLimit} \}) \\
    b_2 &:= (\textit{Decelerate-To-Halt}, \{ \textit{Stop}_\textit{AtEndOfLane} := \textit{True} \})\\
    b_3 &:= (\textit{Decelerate-To-Halt}, \{ \textit{Stop}_\textit{AtStopLine} := \textit{True} \})
\end{align*}

We then define $\mathcal{R}_\textit{man}$ using $b_1,b_2$ and $b_3$:
\begin{align*}
    \mathcal{R}_\textit{man} := \{
    (\{ & True \}, & b_1 )\\
    (\{ & \textit{Ego}_\textit{Approaching} = \textit{Intersection},
    \textit{Crosswalk}_\textit{Obstructed} = \textit{True}  \}, & b_2 )\\
    (\{ & \textit{Ego}_\textit{At} = \textit{Intersection},
    \textit{Crosswalk}_\textit{Obstructed} = \textit{True}  \}, & b_2 )\\
    (\{ & \textit{Ego}_\textit{Approaching} = \textit{Intersection},
    \textit{Road}_\textit{HasStopLine} = \textit{True}  \}, & b_3 )\\
    (\{ & \textit{Ego}_\textit{At} = \textit{Intersection},
    \textit{Road}_\textit{HasStopLine} = \textit{True}\}, & b_3 ) \}\hspace{-0.5em}
\end{align*}

We also define a set of relevant behaviours for the parameter layer, with parameters appropriate for the output of the rule engine:
\begin{align*}
    b_4 &:= (\textit{Track-Speed}, \{  \textit{Ego}_\textit{Speed} := \textit{Target}_\textit{Speed} \})\\
    b_5 &:= (\textit{Decelerate-To-Halt}, \{ \textit{Ego}_\textit{StopAt} := \textit{EndOfLane} \})\\
     b_6 &:= (\textit{Decelerate-To-Halt}, \{ \textit{Ego}_\textit{StopAt} := \textit{StopLine} \})
\end{align*}

Using $b_4,b_5$ and $b_6$, we thus define $\mathcal{R}_\textit{par}$: 
\begin{align*}
    \mathcal{R}_\textit{par} := \{
    (\{ & \textit{Maneuver}_\textit{Track-Speed} = \textit{True} \}, & b_4 )\\
    (\{ & \textit{Maneuver}_\textit{Decelerate-To-Halt} = \textit{True},
    \textit{Stop}_\textit{AtEndOfLane} = \textit{True},\\
    & \hspace{13.2em}\textit{Stop}_\textit{AtStopLine} = \textit{undefined}  \}, & b_5 )\\
    (\{ & \textit{Maneuver}_\textit{Decelerate-To-Halt} = \textit{True},
    \textit{Stop}_\textit{AtStopLine} = \textit{True}  \}, & b_6 ) \}\hspace{-0.5em}
\end{align*}

\sloppy Given the above definition of $\mathcal{R}_\textit{man}$, it follows from \eqref{eq:theory-function} that $\mathcal{F_R}_\textit{man}(s) = \{b_1, b_2, b_3\}$.
The conservativeness of the high-level maneuvers is such that $\textit{Decelerate-To-Halt} \succ \textit{Track-Speed}$, hence $\lambda_\textit{man}(\mathcal{F_R}_\textit{man}(s))$ in \eqref{eq:lambda-man}  gives us: 
\[
\lambda_\textit{man}(\cdot) = (\textit{Decelerate-To-Halt}, \textit{Stop}_\textit{AtEndOfLane} := \textit{True}, \textit{Stop}_\textit{AtStopLine} := \textit{True} \})    
\]

This output is then transformed to the input format of the parameter layer using~\eqref{eq:transformation-par}:
\begin{align*}
    \mathcal{T}_\textit{par}(s) := \{ & \textit{Maneuver}_\textit{Decelerate-To-Halt} := \textit{True},
    && \textit{Stop}_\textit{AtEndOfLane} := \textit{True},\\ 
    & \textit{Maneuver}_\textit{Track-Speed} := \textit{undefined},
    && \textit{Stop}_\textit{AtStopLine} := \textit{True},\\
    & && \textit{Target}_\textit{Speed} := \textit{undefined}\,\}
\end{align*}

With our definition of $\mathcal{R}_\textit{par}$ for the parameter layer, \eqref{eq:theory-function} determines the parameter of the selected high-level maneuver, giving us $\mathcal{F_R}_\textit{par}(\mathcal{T}_\textit{par}(s)) = \{ b_6 \}$. The final output of the rule engine is resolved by \eqref{eq:lambda-par}, giving $\lambda_\textit{par}(\mathcal{F_R}_\textit{par}(\mathcal{T}_\textit{par}(s))) = b_6$.

\section{Learning and Maintaining the Theory}
\label{sec:learning}

To learn the theory, we assume an expert provides a finite set of training scenes $\mathcal{E}\subseteq\mathcal{S}$ and an associated labelling function $\mathcal{L} : \mathcal{E} \to \mathcal{B}$ that assigns a behaviour to every training scene.
Given the characteristics of sets and functions, we know that every scene (a complete set of properties) is unique and is associated to exactly one behaviour.
Since a property may be trivially converted to a constraint using equality, it follows that there always exists a set of rules that can correctly label every scene.

To facilitate learning, we define a backward-chaining \emph{coverage function}
\[
\Phi(r,\mathcal{R},\lambda,\mathcal{T},\mathcal{E}) :=
\{ e \in \mathcal{E}\mid \mathcal{F}_r(\mathcal{T}(e)) \neq \varnothing, \mathcal{F}_r(\mathcal{T}(e)) \in \lambda ( \mathcal{F}_{\mathcal{R}}(\mathcal{T}(e)))\},
\]
which returns the subset of training scenes that \emph{trigger} rule $r$, i.e., cause the rule to contribute to the resolved result of its theory $\mathcal{R}$, associated to a layer with the corresponding resolution function $\lambda\in\{\lambda_\textit{man},\lambda_\textit{par}\}$ and property transformation function $\mathcal{T}\in\{\mathcal{T}_\textit{man},\mathcal{T}_\textit{par}\}$.
The use of the property transformation function allows the training of any layer to be performed with training scenes that are defined w.r.t. the input of the rule engine.
$\mathcal{T}_\textit{par}$ is given in~\eqref{eq:transformation-par}. The maneuver layer requires no transformation, so $\mathcal{T}_\textit{man}$ is simply the identity transformation,
\begin{equation*}
\mathcal{T}_\textit{man}(e\in\mathcal{S}) := e.
\end{equation*}

\subsection{Rule Engine Update Algorithm}
\label{SS:RuleEngineAlgorithms}

The main method of the Rule Engine Update
algorithm (\cref{alg:RuleEngineRefinement}) exploits the common structure of the two layers, calling the RuleUpdate subroutine (\cref{alg:RuleRefinement}) per layer. Since the rules of the parameter layer are dependent on those of the maneuver layer, \cref{alg:RuleRefinement} is called on the maneuver layer first.

\begin{algorithm}
\LinesNotNumbered
\setstretch{1.1}
\SetKwInOut{Input}{input}
\caption{Rule Engine Update}
\label{alg:RuleEngineRefinement}
\Input{
\begin{tabular}[t]{ll@{\quad}ll}
$\mathcal{E}$ & training scenes &
$\mathcal{R}_\textit{man}$ & base maneuver rules\\
$\mathcal{L}$ & labelling function &
$\mathcal{R}_\textit{par}$ & base parameter rules
\end{tabular}
}
\vspace{6pt}
$\mathcal{R}_\textit{man} \leftarrow \text{RuleUpdate}\left(\mathcal{E}, \mathcal{L}, \mathcal{R}_\textit{man}, \lambda_\textit{man}, \mathcal{T}_\textit{man}\right)$\\
$\mathcal{R}_\textit{par} \leftarrow \text{RuleUpdate}\left(\mathcal{E}, \mathcal{L}, \mathcal{R}_\textit{par}, \lambda_\textit{par}, \mathcal{T}_\textit{par}\right)$\\
\Return $\left(\mathcal{R}_\textit{man}, \mathcal{R}_\textit{par} \right)$
\end{algorithm}

The rule engine works by filtering a set of candidate behaviours.
The purpose of~\cref{alg:RuleRefinement} is to modify or create rules such that the set of behaviours output by a layer contains the correct behaviour for every training scene.
Other than in unusual pathological cases (described below), the algorithm will find a theory that satisfies this requirement.
The following description applies to either layer.

\Cref{alg:RuleRefinement} is given an existing theory $\mathcal{R}$ that may be empty---the algorithm will generate any new rules it needs. 
In \cref{alg:start}, the algorithm initializes an empty set of bad rules $\mathcal{R}_\textit{Bad}$.
This set is used to contain any rules that are discovered to have no coverage in the training scenes. Such rules may already exist in $\mathcal{R}$ or may be generated as candidates by the algorithm.

\begin{algorithm}
\SetKwInOut{Input}{input}
\caption{Rule Update}
\label{alg:RuleRefinement}
\Input{
\begin{tabular}[t]{ll@{\quad}ll}
$\mathcal{E}$ & training scenes & $\lambda$ & resolution function\\
$\mathcal{L}$ & labelling function &
$\mathcal{T}$ & property transformation function\\
$\mathcal{R}$ & base rule set\\
\end{tabular}
}
$\mathcal{R}_\textit{Bad} \leftarrow \varnothing$\label{alg:start}\\
\While{$\left( \varepsilon \leftarrow \left\{ e \in \mathcal{E} \mid \mathcal{L}(e) \notin \lambda\left( \mathcal{F}_{\mathcal{R}}\left(\mathcal{T}(e)\right) \right) \right\} \right) \neq \varnothing$\label{alg:while}}{
Select randomly $e \in \varepsilon$\label{alg:e-in-epsilon}\\
\If{$\nexists r \in \mathcal{R} \mid \mathcal{F}_r\left(\mathcal{T}(e)\right) = \mathcal{L}(e)$\label{alg:if-start}}{
    $\mathcal{R} \leftarrow \mathcal{R} \cup \{(\textit{True}, \mathcal{L}(e)) \}$\\
    \label{alg:if-end}
}
\Else{\label{alg:else}
Select randomly $r := (\textit{antecedent}, \textit{consequent}) \in$ \\
\hspace{\fill} $\left\{ r \in \mathcal{R} \mid \mathcal{F}_r\left(\mathcal{T}(e)\right) \in \lambda \left( \mathcal{F}_{\mathcal{R}}\left(\mathcal{T}(e)\right)\right)\right\} \label{alg:r-in-delta}$\\
$K \leftarrow \bigcup\Phi \left(r, \mathcal{R},\lambda,\mathcal{T},\mathcal{E} \right)$\label{alg:K-gets-Phi}\\
$C \gets \textit{GenerateConstraints}\,(K,\{=,\leq,\geq\})$\label{alg:C-gets}\\
$\mathcal{R} \leftarrow \mathcal{R} \setminus \{ r \}$\label{alg:R-gets}\\
\Repeat{$r' \notin \mathcal{R}_\textit{Bad} \cup \mathcal{R}$\label{alg:until}}{\label{alg:repeat}
\If{$C = \varnothing$\label{alg:repeat-if}}{
    \textbf{throw}
    \textit{BadBaseRules}
    \label{alg:throw}
}
$c \gets \textit{GetConstraint}\,( C,r,e,\mathcal{E},\mathcal{L},\mathcal{R},\lambda,\mathcal{T})$ \label{alg:c-in-C}\\
$C \leftarrow C \setminus \{ c \}$\label{alg:K-gets-K}\\
$r' \gets (\textit{antecedent}\textsf{ AND } c, \textit{consequent})$\label{alg:r-prime-gets}
}
\If{$\nexists e \in \mathcal{E} \mid \mathcal{F}_{r'}\left(\mathcal{T}(e)\right) = \textit{consequent}$\label{alg:until-if}}{
    $\mathcal{R}_\textit{Bad} \leftarrow \mathcal{R}_\textit{Bad} \cup \{ r' \}$
    \label{alg:R-Bad-gets}
}
\Else{\label{alg:until-else}
    $\mathcal{R} \leftarrow \mathcal{R} \cup \{ r' \}$
    \label{alg:while-end}
}
}
}
\Return $\mathcal{R}$\label{alg:return}
\end{algorithm}

The outer loop of the algorithm is controlled by the existence of training scenes that are misclassified, i.e., when the set of output behaviours of the layer does not contain the specified label of the scene (\cref{alg:while}).
If there are no misclassifications, the algorithm terminates correctly by returning the current theory in~\cref{alg:return}.
If there exist misclassified scenes, one is selected at random in~\cref{alg:e-in-epsilon}.
We use random selection to avoid giving undue bias to any particular solution.
If there is no rule whose consequent is the labelled behaviour with the chosen scene, the most general rule for this behaviour is added to $\mathcal{R}$ in~\cref{alg:if-start,alg:if-end}.
The main rule-generating section of the algorithm then follows.

A rule $r$ that is triggered by the chosen misclassified scene is selected at random in~\cref{alg:r-in-delta}.
Once again, random selection is used to avoid bias.
\Cref{alg:K-gets-Phi} generates the set of properties $K$, containing all the properties in the scenes that trigger $r$.
\Cref{alg:C-gets} then creates a set of feasible constraints $C$, given $K$ and the operators $\{=,\leq,\geq\}$.
Each of these operators includes equality to ensure that every constraint covers the property observed in a training scene.
The chosen rule $r$ is then removed from the current theory $\mathcal{R}$ in~\cref{alg:R-gets}.
This allows $r$ to be updated and re-inserted or rejected if the update turns out to be bad, i.e., have no coverage in the training scenes.

The repeat loop in~\crefrange{alg:repeat}{alg:until} creates a novel candidate rule $r'$ by adding a single constraint $c$ to the antecedent of $r$ in~\cref{alg:r-prime-gets}.
In~\cref{alg:c-in-C}, function \textit{GetConstraint} chooses $c$ from $C$ according to a heuristic criterion that aims to improve the chance that the misclassification will eventually be resolved, such as precision, coverage difference, rate difference, or Laplace estimate~\cite{furnkranz2012foundations}.
This implies that the new constraint will not conflict with the existing ones.
A single additional constraint may not resolve the misclassification, but it is sufficient for the algorithm's correct termination that the candidate rule is novel w.r.t. the union of $\mathcal{R}_\textit{Bad}$ and $\mathcal{R}$ (\cref{alg:until}).
Informally, the existence of a novel candidate is ensured by the fact that every training scene is uniquely defined by its properties---which may be trivially converted into constraints---and that constraints may be added to a rule until it specifies a unique scene.
A counterexample to this intuition is the unusual pathological case when a 
rule in the base rule set is triggered by a misclassified example and already contains all the constraints derivable from the training scenes.
Under these circumstances, the repeat loop will exhaust all candidates and throw an exception (\cref{alg:repeat-if,alg:throw}).
In such a case, the aberrant rule must be removed from the base set.

On exiting the repeat loop, the candidate rule $r'$ is guaranteed novel, but may not be good.
If $r'$ has no coverage in the training scenes, \cref{alg:R-Bad-gets} adds it to $\mathcal{R}_\textit{Bad}$.
If there is coverage, $r'$ is added to $\mathcal{R}$, although this does not guarantee that it will immediately resolve the misclassification.
This is achieved by the repeated checking and iteration provided by the outer loop, and by the non-zero chance that $r'$ will be further refined, if this is necessary.
The antecedents of rules thus increase monotonically, becoming more specific until the point at which they resolve a misclassification, or have no coverage and are rejected.

\Cref{alg:RuleRefinement} does not necessarily converge monotonically: adding a constraint to a rule makes it more specific, potentially increasing the number of misclassifications observed in~\cref{alg:while} before all necessary updates are completed. Removing bad candidate rules from $\mathcal{R}$ in~\cref{alg:R-Bad-gets} leaves the theory temporarily incomplete and has the same effect.
The number of bad rules in $\mathcal{R}_\textit{Bad}$ does increase monotonically, thus ensuring eventual termination.
In the worst case, the algorithm will suffer the exponential complexity of trying all rules, with all possible combinations of constraints.
This does not happen in practice because function \textit{GetConstraint} in~\cref{alg:c-in-C} avoids obvious conflicts and makes heuristically good choices.
\subsection{Rule and Training Set Development}
\label{S:RuleSetDevelopment}

\begin{wrapfigure}{r}{0.55\textwidth}
    \vspace{-2em}
    \centering
    \begin{tikzpicture}
    \textsf{\scriptsize
    \draw[every node/.style={draw,fill=blue!10},text width=7em,text depth=1.3em,align=center]
    (0,0)node(DI){Discrepancy Identification}
    (4,0)node(MD){Misbehaviour Diagnosis}
    (4,-2)node(KE){Knowledge Extraction}
    (0,-2)node(RE){Rule\\Engineering};
    \draw[->,align=center,text width=6em]
    (DI)edgenode[above]{Discrepancy Scene}(MD)
    (MD)edgenode[right,align=left]{Conflicting Scenes}(KE)
    (KE)edgenode[below]{Discriminating Scene}(RE)
    (RE)edgenode[left,align=right,text width=4em]{Satisfying Rules}(DI);
    }
    \end{tikzpicture}
    \caption{Knowledge engineering cycle}
    \label{F:KnowledgeEngineeringMethod}
    \vspace{-2em}
\end{wrapfigure}
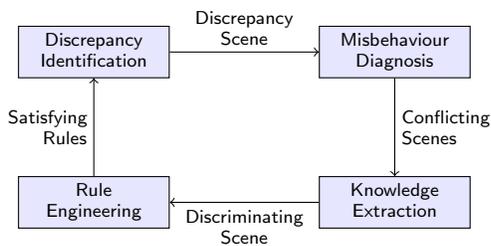

Our algorithms are sufficient to find a rule-based theory that perfectly agrees with a set of labelled training scenes; however, they do not guarantee the understandability of the theory's decisions.
To bridge this gap between theory and practice, we give here an outline of a knowledge engineering cycle that allows an expert to incrementally build a set of discriminating training scenes and corresponding rules.
The four steps of the cycle, illustrated in~\cref{F:KnowledgeEngineeringMethod}, are described below.

\subsubsection{Discrepancy Identification}
\label{sec:discrepancy}
Rule set development is prompted by the existence of discrepancies between the actual and desired behaviour of the rule engine.
These usually occur when the rule engine encounters a novel scene during deployment.
Hence, the first step is to identify a scene that exemplifies the discrepancy, either from test suites, simulation testing, recordings of traffic flow, or open-road testing.

\subsubsection{Misbehaviour Diagnosis}
\label{sec:diagnosis}
A discrepancy is not necessarily a misbehaviour of the rule-based theory.
To determine whether it is the fault of the theory or something external, such as perception noise, forward and backward chaining are used to identify the training scenes that also trigger the rules that misclassify the discrepancy scene. We call these scenes the conflicting scenes, since they are similar to the discrepancy scene but their behaviour label is different from the desired label for the discrepancy scene.
This procedure is in line with how \cref{alg:RuleRefinement} works. The comparison of the discrepancy and conflicting scenes helps understand the discrepancy. 
If the analysis reveals that the current decision for the discrepancy scene is reasonable as is, or that the problem is external to the rule-based theory, there is no need to proceed.
Otherwise, the theory is deemed incomplete and the conflicting scenes, including the discrepancy scene, are passed to the next step.

\subsubsection{Knowledge Extraction}
\label{sec:acquisition}
The discrepancy scene along with its correct decision is an example of behaviour that must be incorporated in the theory; however, it may contain irrelevant properties that make it too specific to be used without modification, such as vehicles in the scene that are irrelevant to concluding the correct behaviour.
Such properties increase the number of existing rules the new scene triggers, potentially causing \cref{alg:RuleRefinement} to produce new rules that are unnecessarily complex and opaque.
The purpose of this step is therefore to analyse the conflicting scenes and eliminate irrelevant properties from the discrepancy scene by setting them as undefined.

\subsubsection{Rule Engineering}
\label{sec:engineering}
Once the discrepancy scene has been sanitized in the previous step, it is added to the existing set of training scenes.
The rule-based theory is then updated using \cref{alg:RuleEngineRefinement}, with input parameters $\mathcal{R}_\textit{man}$ and $\mathcal{R}_\textit{par}$ set to the existing theory. Assuming that this theory correctly classifies the previous training set (certainly true if it was produced by \cref{alg:RuleEngineRefinement}), the outer loop of \cref{alg:RuleRefinement} will necessarily select the discrepancy scene as the first misclassification to repair. The algorithm will then incrementally refine the rules, as previously described, until every training scene derives its expected behaviour.

Incorporating the sanitized discrepancy scene in the rule-based theory does not guarantee that the original misbehaviour will be cured, not even for the specific instance.
The new theory is thus re-evaluated and the knowledge engineering cycle is repeated, until no further discrepancies are detected.
\section{Experimental Results}
\label{sec:results}

Using the schema outlined in~\cref{sec:rule-engine} as a guide, we developed a prototype rule engine in ECMAScript 2016-2017~\cite{ecmascript} (standardized JavaScript), using polyfills to ensure consistent behaviour with different interpreters.
Our prototype makes use of many optimizations not described in the text, including the use of quantifiers over constraints in the syntax of rule antecedents, and caching the results of rule antecedent evaluation (memoization).
We executed the rule engine on Google's V8 interpreter\footnote{https://V8.dev} in all the experiments described below. 

The software stack of our autonomous vehicle is based on the Robot Operating System (ROS), within which the behaviour planner is a ROS node written in C++.
The behaviour planner node processes the autonomous vehicle's sensor data and communicates with the rule engine via JSON streams, using the RESTful application programming interface.

\subsection{Driving Policy}

To learn the sets of maneuver and parameter rules of our rule engine, we incrementally built a test suite consisting of 683 labelled scenes. Each scene was expertly curated following the method presented in~\cref{S:RuleSetDevelopment}.
Using this test suite, we constructed a rule-based theory consisting of 330 maneuver rules and 16 parameter rules. The distributions of maneuver and parameter rules are shown in \cref{F:RuleDistribution,F:ParameterRuleDistribution}, respectively. 
From \cref{F:RuleDistribution}, we see that $111/330\approx33.6\%$ of the maneuver rules enforce an \textit{Emergency-Stop}, delimiting the rule engine's operational design domain (ODD). However, we note that 63 of these rules are only required to ensure that the environment representation is well-formed and that its attributes are used coherently during software integration. These rules can therefore be removed once the integration is complete, and we may reasonably conclude that the autonomous vehicle can drive in an urban environment with only 267 maneuver rules.
\Cref{F:ParameterRuleDistribution} illustrates that the number of parameter rules for a given high-level maneuver reflects the number of different ways in which the maneuver is used.  

\begin{figure}[ht!]
\begin{minipage}{0.49\textwidth}
\begin{tikzpicture}
\sf
\pgfplotsset{%
    width=0.95\textwidth,
    height=12em
}
\begin{axis}[ 
    style={font=\scriptsize},
    xbar,
    axis line style={draw=none},
    tick style={draw=none},
    xmin=0,
    xlabel={Number of maneuver rules},
    xticklabels=\empty,
    symbolic y coords={{Emergency-Stop}, {Decelerate-To-Halt}, {Follow-Leader}, {Track-Speed}, {Yield}, {Pass-Obstacle}, {Stop}},
    ytick=data,
    nodes near coords,
    % every node near coord/.append style={anchor=east}
]
\addplot[fill=blue!20,draw=none] coordinates {(111,{Emergency-Stop}) (67,{Decelerate-To-Halt}) (54,{Follow-Leader}) (53,{Track-Speed}) (20,{Pass-Obstacle}) (23,{Yield}) (2,{Stop})};
\end{axis}
\end{tikzpicture}
\caption{Maneuver rule distribution}
\label{F:RuleDistribution}
\end{minipage}
\qquad
\begin{minipage}{0.48\textwidth}
\begin{tikzpicture}
\sf
\pgfplotsset{%
    width=0.75\textwidth,
    height=12em
}
\begin{axis}[ 
    style={font=\scriptsize},
    xbar,
    axis line style={draw=none},
    tick style={draw=none},
    xmin=0,
    xlabel={Number of parameter rules},
    xticklabels=\empty,
    symbolic y coords={{Emergency-Stop}, {Decelerate-To-Halt}, {Follow-Leader}, {Track-Speed}, {Yield}, {Pass-Obstacle}, {Stop}},
    ytick=data,
    nodes near coords
]
\addplot[fill=blue!20,draw=none] coordinates {(0,{Emergency-Stop}) (8,{Decelerate-To-Halt}) (3,{Follow-Leader}) (1,{Track-Speed}) (2,{Pass-Obstacle}) (2,{Yield}) (0,{Stop})};
\end{axis}
\end{tikzpicture}
\caption{Parameter rule distribution}
\label{F:ParameterRuleDistribution}
\end{minipage}
\end{figure}

\subsection{Field Test}

To demonstrate the viability of the rule-based theory within its ODD, the rule engine was deployed in the University of Waterloo autonomous vehicle (shown in~\cref{fig:moose-motion-planner}) and tested by driving 110\,km on public roads in full autonomy.
During the public drive, the rule engine was able to serve queries at up to 300\,Hz. Given that the typical rate of other components in our autonomous vehicle's software stack is 10\,Hz, the rule-based theory was far from being the bottleneck.

The public drive was performed on a network of two-lane commercial roads constrained by four-way intersections and T-intersections, with precedence varying between the autonomous vehicle (AV) and other dynamic objects (DO).
The route was planned to give the autonomous vehicle a non-trivial driving challenge. For instance, it had to effectively handle unprotected left-hand turns and avoid a myriad of parked vehicles. In this latter case, the rules were implemented to ensure that the autonomous vehicle could safely pass the parked vehicles by temporarily encroaching on the oncoming lane.
The numbers of various road elements encountered and the numbers of different behaviours performed by our autonomous vehicle are as follows:

\begin{center}
\begin{tabular}{rl@{~}|@{~}rl}
48 & Four-way intersections, precedence for AV &
120 & Straight crossings\\
60 & Four-way intersections, precedence for DO &
36 & Protected left turns\\
144 & T-intersections, precedence for AV &
84 & Unprotected left turns\\
72 & T-intersections, with precedence for DO &
84 & Right turns\\
24 & Cul-de-sacs
\end{tabular}
\end{center}

During the public drive, a safety driver had to intervene 58 times, with an average time between interventions of over 5 minutes and 30 seconds.
This is comparable to the reported performance of the end-to-end deep-learning approach of~\cite{bojarski2016end}.
Using metrics defined in that work, both approaches achieve \textapprox98\% autonomy, albeit with different ODDs.
In our case, we believe that many of the interventions derive from limitations external to the rule engine and could be avoided.
Approximately 64\% of the interventions were due to the driving scenario encountered being out of the ODD. Most of these scenarios involved interaction with dynamic objects that our prototype system was not programmed to detect, such as animals, bicycles, three-wheelers, and heavy vehicles. Approximately 36\% of the interventions  were due to the cautious design of the driving policy, resulting in deadlocks at some intersections when the dynamic object tracker was unable to determine whether a vehicle was parked or was slowly moving (and thus might have precedence to enter in the intersection).

\section{Conclusion}
\label{sec:conclusion}
We have defined a two-layer rule engine and provided an algorithm to create and maintain its rule-based theory. We have demonstrated the practicality of our approach by constructing a prototype that has been used to drive our level-3 autonomous vehicle more than 110\,km in a busy urban environment.
Our rule engine required few human interventions, achieving a similar degree of autonomy to that of a highly-cited state-of-the-art approach based on deep learning~\cite{bojarski2016end}.

Our ongoing work is focused on extending the operational design domain (ODD) of our prototype rule engine, adding new high-level maneuvers and further stratifying its rule-based theory to handle more complex driving scenarios.
Our current learning algorithm constructs a theory that perfectly agrees with the training scenes, or reports the inconsistency.
To accommodate imperfect training data, which may contain unavoidable inconsistencies, we are also extending our algorithm to refine rules based on statistical performance.
The resulting theory may then be more robust to inconsistencies encountered during deployment.

Although we encountered no problems of tractability in constructing our prototype, we realize that this might not be the case if there are significantly more rules or if the data structures are significantly enlarged.
On the other hand, it is not clear that our rule-based approach will be less tractable than standard machine-learning approaches, which are known to be data inefficient.
To investigate this, we have incorporated the two-layer rule engine into our reinforcement learning platform for autonomous driving~\cite{Lee2019}.
This will allow us to make a direct comparison of tractability using the same ODD, and also to investigate how the rule engine performs when other vehicles drive unpredictably or adversarially.

\section*{Acknowledgment}
This work was supported throughout its development by Fonds de Recherche du Qu\'{e}bec – Nature et Technologies (FRQNT) and Natural Sciences and Engineering Research Council (NSERC) Discovery Grant. Author SS was supported by Japanese
Science and Technology agency (JST) ERATO project JPMJER1603: HASUO Metamathematics for Systems Design.

\bibliographystyle{splncs04}
\bibliography{bibliography}

\end{document}